# Translating OWL and Semantic Web Rules into Prolog: Moving Toward Description Logic Programs

Ken Samuel[1], Leo Obrst[1],
Suzette Stoutenberg[2], Karen Fox[2], Paul Franklin[2], Adrian Johnson[2],
Ken Laskey[1], Deborah Nichols[1], Steve Lopez[2], Jason Peterson[2]
{samuel, lobrst, suzette, kfox, pfranklin, abjohnson,
klaskey, dlnichols, slopez, jasonp}@mitre.org

The MITRE Corporation
[1] 7525 Colshire Drive, McLean, VA 22102-7508
[2] 1155 Academy Park Loop, Colorado Springs, CO 80910-3716

**Abstract.** We are researching the interaction between the rule and the ontology layers of the Semantic Web, by comparing two options: 1) using OWL and its rule extension SWRL to develop an integrated ontology/rule language, and 2) layering rules on top of an ontology with RuleML and OWL. Toward this end, we are developing the SWORIER system, which enables efficient automated reasoning on ontologies and rules, by translating all of them into Prolog and adding a set of general rules that properly capture the semantics of OWL. We have also enabled the user to make dynamic changes on the fly, at run time. This work addresses several of the concerns expressed in previous work, such as negation, complementary classes, disjunctive heads, and cardinality, and it discusses alternative approaches for dealing with inconsistencies in the knowledge base. In addition, for efficiency, we implemented techniques called extensionalization, avoiding reanalysis, and code minimization.

## 1. Introduction

The wedding of Semantic Web technology and Logic Programming has created a new technical paradigm called *description logic programming*. Recently, researchers have begun focusing on the question of how to utilize logical rules from a particular domain in order to improve the Semantic Web (Bechhofer *et al.* 2004; Hirtle *et al.* 2004; Horrocks *et al.* 2004). It is well established that the knowledge development language and the knowledge runtime language may not be the same, so the former should be transformed to the latter for greater efficiency at run time (Cadoli & Donini 1997; Darwiche & Marquis 2001).

We have developed SWORIER (Semantic Web Ontologies and Rules for Interoperability with Efficient Reasoning), which is a system that uses Logic Programming to reason about and answer queries about ontologies and rules (Grosof *et al.* 2003; Hitzler *et al.* 2005; Volz *et al.* 2003). OWL (Web Ontology Language) ontologies (Bechhofer *et al.* 2004), along with rules in the Semantic Web Rule Language (SWRL) (Horrocks *et al.* 2004) or the Rule Markup Language (RuleML) (Hirtle *et al.* 2004) are all translated into Prolog using XSLTs (Extensible Stylesheet Language Transformations). In addition, we have written a set of *General Rules* in Prolog in order to enforce the semantics of OWL primitives. To do this, it was necessary to address a number of issues related to negation, the open world assumption, complementary and disjoint classes, disjunctive conclusions, enumerated classes, equivalent individuals, error messages, existential quantification, cardinality constraints, duplicate facts, cyclical hierarchies, and anonymous classes. Recent work has suggested that some of these problems are unsolvable (Volz *et al.* 2003), but we believe we have found solutions for them.

We have imposed strong efficiency requirements, demanding that queries are answered in a matter of seconds. And, unlike previous work, SWORIER can assimilate dynamic changes that are provided at run time, including adding new facts, removing facts, and swapping rule sets, which also must be done in seconds. To achieve this level of efficiency, we established three techniques: extensionalization, avoiding reanalysis, and code minimization.

This paper is organized as follows: First, Section 2 provides the reader with background information. Then, Section 3 describes SWORIER's system design. Section 4 addresses the challenges found in previous research. Section 5 adds a capability to handle dynamic changes. Section 6 analyzes and addresses the efficiency of the system, and then Section 7 discusses related work. Finally, Section 8 summarizes the paper, offers conclusions, and discusses future work.

## 2. Background

We are investigating the interaction between rules and ontologies in the Semantic Web to determine how a standard language should best express them (Stoutenburg *et al.* 2006). In particular, to determine whether an ontology and the corresponding rules should be integrated or layered, we specified, translated, and executed information in: 1) SWRL, which integrates OWL with rules (Horrocks *et al.* 2004), and 2) RuleML layered on top of OWL. The ontologies and instances were developed in the Cerebra OWL ontology development environment,[1] the SWRL and RuleML rules were created in a text editor, and AMZI! (2006) Prolog was used as the inference engine. We translated the constructs for both the integrated and layered approaches into Prolog code, gauging each approach in terms of effectiveness, efficiency, difficulty, restrictiveness, translatability, and suitability for deployment in an operational setting.

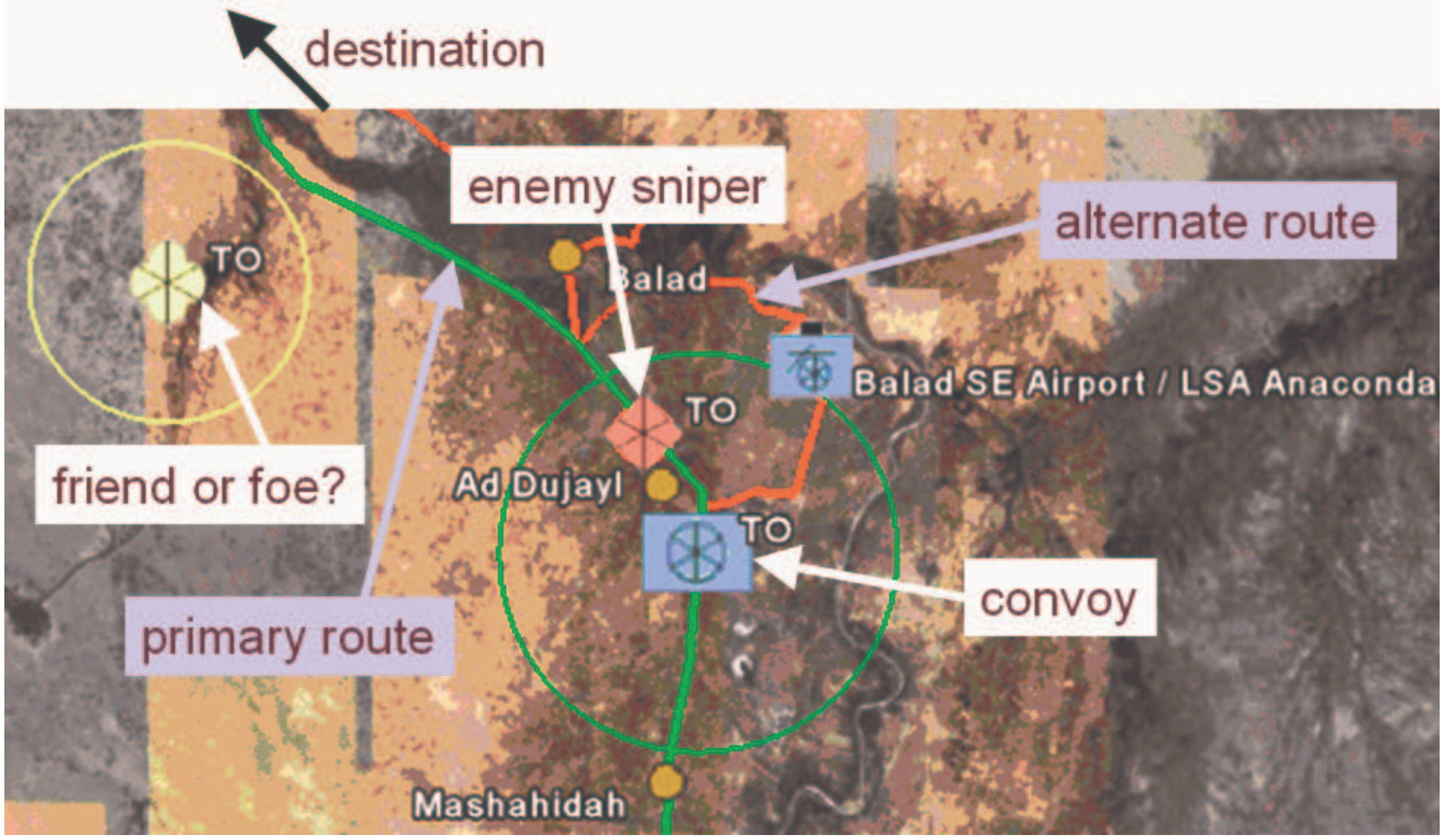

**Figure 1. A Military Task**

[1] http://cerebra.com/index.html

Our initial experiments were set in a military command and control domain with a supply convoy moving through an unsecured area. Figure 1 presents an example situation, where a convoy is moving north along the primary route, approaching the location where intelligence has reported an enemy sniper is stationed. New information can become available at any time, such as the discovery of a theater object or the beginning of a sandstorm. The system has rules that trigger alerts and recommendations to report to the convoy commander. For example, in the situation shown in Figure 1, an enemy unit is within the convoy's region of interest (the circle surrounding the convoy), so the system might tell the convoy commander, "ALERT: Intelligence report of enemy sniper in the vicinity." and "RECOMMENDATION: Take alternate route."

Most knowledge representation languages and knowledge-based systems utilize a restricted version of First Order Logic (FOL). FOL, however, is semi-decidable. It is decidable in that if a theorem is logically entailed by a FOL theory, a proof will eventually be found, but it is undecidable in that if a theorem is not logically entailed, a proof of that may never be found. But decidability here does not mean tractability, and in general even inference in the simpler propositional calculus is NP-complete (Cadoli *et al.* 1999), i.e., usually unable to be processed in less than exponential time.

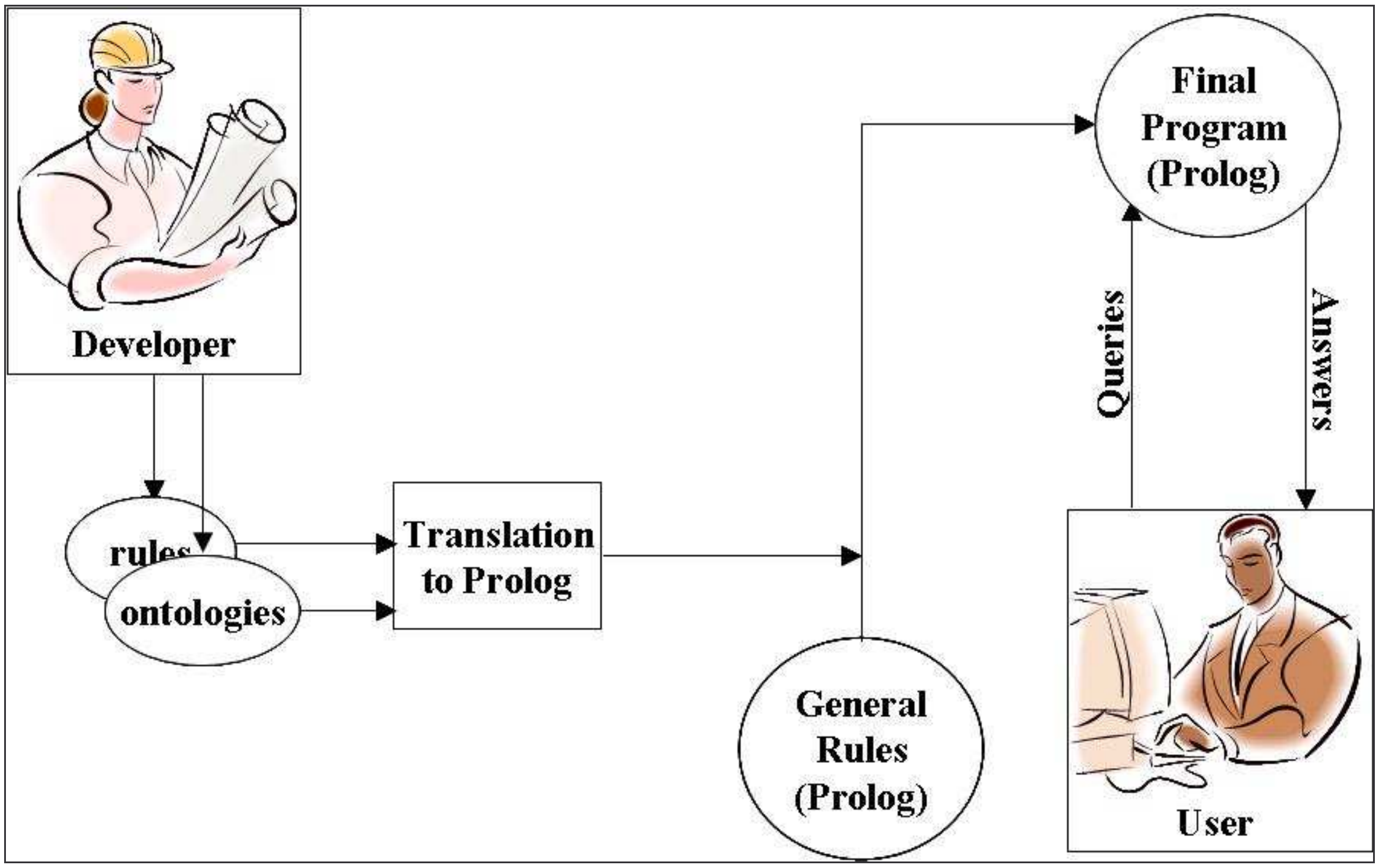

**Figure 2. System Design**

To make inference tractable, various approaches in the field of *knowledge compilation*, which involves converting a knowledge base into a more concise or tractable representation, have been devised (Cadoli & Donini 1997; Darwiche & Marquis 2001). One approach is to syntactically restrict the knowledge representation language, sacrificing expressiveness for tractability and efficiency (de Bruijn *et al.* 2004). Logic programming (LP), description logic (DL), and description logic programming (an emerging field that weds DL and LP) take this approach (Ait-Kaci 1991; Grosof *et al.* 2003; Hitzler *et al.* 2005; Van Roy 1990; Volz *et al.* 2003). For example, OWL is a DL that defines a tractable subset of First-Order Logic (Bechhofer *et al.* 2004; Daconta *et al.* 2003). An alternative is to employ theory approximation (Kautz &

Selman 1994; Kautz & Selman 1991), in which the queries that are logically entailed by a knowledge base (a “theory”) can be correctly answered, while, for the rest of the queries, the response is “unknown”. Some researchers preprocess the knowledge in various ways to relax either the completeness or the soundness requirements, perhaps by generating certain default conclusions (Cadoli & Donini 1997). Another possible optimization is to extensionalize the rule base, as discussed in Section 6.1.

## 3. System Design

Figure 2 shows the system design of SWORIER. A *developer* creates ontologies, knowledge bases, and/or rules in the formalism(s) of OWL, RuleML, and/or SWRL. Examples of OWL, RuleML, and SWRL are in Table 1a, 1b, and.1c, respectively. This information is translated into Prolog code using XSLTs, resulting in the code shown in Table 2a, 2b, and 2c. (We include words like "is" and "of" in our predicate names to avoid ambiguity. Otherwise, there is the danger of misinterpreting the roles of the arguments. For example, `member(X, Y)` could be interpreted as "X is a member of Y" or "X has a member, Y", while `ismemberof(X, Y)` is more clear.) Finally, a set of *General Rules* (defined in Section 3.2) is appended to the XSLT output to form a complete Prolog program, which can be queried by the *user*.

**Table 1. Examples of OWL, RuleML, and SWRL**

| | |
|---|---|
| a. | <sniper rdf:ID="smith"><br> <hasCombatIntent rdf:resource="#friendlyIntent"/><br></sniper> |
| b. | <Implies><br> <head> <Atom><br>  <opr> <Rel>redForceTheaterObject</Rel> </opr><br>  <Var>X</Var><br> </Atom> </head><br> <body> <Atom><br>  <opr> <Rel>redForceTheaterObject</Rel> </opr><br>  <Var>X</Var><br> </Atom> </body><br></Implies> |
| c. | <swrlx:classAtom><br> <owlx:Class owlx:name = "sniper"/><br> <owlx:Individual owlx:name="smith"/><br></swrlx:classAtom><br><ruleml:imp><br> <ruleml:_body><br>  <swrlx:individualPropertyAtom swrlx:property="isDescribedBy"><br>   <ruleml:var>T</ruleml:var><br>   <ruleml:var>G</ruleml:var><br>  </swrlx:individualPropertyAtom><br>  <swrlx:individualPropertyAtom swrlx:property="hasSpeedObservation"><br>   <ruleml:var>G</ruleml:var><br>   <ruleml:var>S</ruleml:var><br>  </swrlx:individualPropertyAtom><br> </ruleml:_body><br> <ruleml:_head><br>  <swrlx:individualPropertyAtom swrlx:property="hasSpeed"><br>   <ruleml:var>T</ruleml:var><br>   <ruleml:var>S</ruleml:var><br>  </swrlx:individualPropertyAtom><br> </ruleml:_head><br></ruleml:imp> |

### 3.1. Translating Facts

SWORIER uses a syntax different from that typically found in previous work. For example, Volz *et al.* (2003) would produce the translation of Table 2d, instead of the translation in Table 2a. But we note that the syntax used by Volz *et al.* (2003) cannot represent "every class that `smith` is a member of" with `X(smith)`, because most Prolog implementations disallow predicate variables. In contrast, by making the class names and property names be arguments instead of predicates, SWORIER has the flexibility to generalize on them with, for example, `ismemberof(smith, X)`.

**Table 2. Translations**

| | |
|---|---|
| a. | `ismemberof(smith, sniper).`<br>`haspropertywith(smith, hasCombatIntent, friendlyIntent).` |
| b. | `ismemberof(X, redForceTheaterObject) :-`<br>` isMemberOf(X, redForceTheaterObject).` |
| c. | `ismemberof(smith, sniper).`<br>`haspropertywith(T, hasSpeed, S) :-`<br>` hasPropertyWith(T, isDescribedBy, G),`<br>` hasPropertyWith(G, hasSpeedObservation, S).` |
| d. | `sniper(smith).`<br>`hasCombatIntent(smith, friendlyIntent).` |

### 3.2. General Rules

The General Rules are meant to capture the semantics of the primitives in OWL. For example, the rules in Table 3a enforce the transitivity of subclass. Note that there are two different predicates: `issubclassof` and `isSubClassOf`. One predicate would be insufficient, because Table 3b has left recursion, resulting in an infinite loop.

**Table 3. The Transitive Closure of Subclass**

| | |
|---|---|
| a. | `isSubClassOf(C, D) :- issubclassof(C, D).`<br>`isSubClassOf(C, E) :- issubclassof(C, D), isSubClassOf(D, E).` |
| b. | `isSubClassOf(C, E) :- isSubClassOf(C, D), isSubClassOf(D, E).` |

With two different subclass predicates, some questions must be answered. Should the user submit queries with `issubclassof` or `isSubClassOf`? Also, which form should the input from the XSLTs be? If the input used `isSubClassOf`, then neither of the rules in Table 3a would ever succeed, thus the input must use `issubclassof`. On the other hand, queries should use `isSubClassOf` because the `issubclassof` set of facts is incomplete —— none of the subclass relationships that are derived by transitivity are captured by `issubclassof`. Note that the `issubclassof` set of facts is a subset of the `isSubClassOf` set of facts, because of the first rule in Table 3a, which is called the *conversion rule* for subclass.

For consistency, we created two cases of each predicate, all-lowercase and camelcase.[2] Also, each predicate has a conversion rule. The XSLT facts always use the all-lowercase forms of predicates, while the user queries are always in camelcase. (However, the developer decides how to spell the names of constants, such as `hasSpeed` in Table 2c.) And any rules, other than recursive rules and conversion rules, follow the convention:

[2] Any predicates that are not used for input or output are written in an underscore case, such as `is_sub_class_of_but_not_equal_to`. Also, for some predicates, there are two sources of recursion, requiring three cases of the predicate. An example of this is the member relation, for which the three cases are `ismemberof`, `is_member_of`, and `isMemberOf`.

*All predicates in the body of the rule are camelcase,*
*and the predicate in its head is all-lowercase.*

As an example, see the second rule in Table 2c. Using camelcase predicates in the rule's body guarantees that the rule is triggered by everything that can be derived for that predicate in either case. And using the all-lowercase predicate for the rule's head insures that any facts generated by the rule will hold for both cases of the predicate.

### 3.3. Translating Rules

Some of the inputs provided to SWORIER are RuleML or SWRL rules that were created by the developer. It is not difficult to translate these rules into Prolog, because they are written in Horn Clause form. However, we cannot control which rules are provided nor how they are written. Problems can emerge, such as the infinite loop in Table 2b, which was generated by the RuleML rule in Table 1b. In addition, the order in which rules are listed and the order of the terms in the rules' bodies can have significant effects, much like order of join evaluation in database languages such as SQL. Another concern is that the input could include rules that produce duplicate copies of a fact. (See Section 4.9.) And the rules might be inefficient.

There are ways to correct or at least mitigate some of these problems. For example, we could apply transformations as for logic queries in the form of rewrite rules such as “magic sets” optimization (Cadoli *et al.* 1999; Sippu & Soisalon-Soininen 1996). But currently, we must impose strong requirements on the developer, who may need to be very familiar with Prolog programming techniques, the logical consequences of the facts in the ontologies, and the General Rules.

## 4. Challenges

We are following the groundbreaking work of Volz *et al.* (2003), who were among the first researchers to investigate OWL-to-Prolog translation. They discussed a number of problems that they encountered in the course of their work. Now we are proposing solutions for several of these problems, some of which are currently implemented in SWORIER.

### 4.1. Negation

Negation in Prolog is not the same as negation in OWL, RuleML, and SWRL. Prolog has *finite-failure negation*, which means that not(T) is true if it is not possible to prove that T is true. Alternatively, with the *logical negation* of OWL, RuleML, and SWRL, not(T) is true if it can be proven that T is false. (In both cases, not(T) is false if T can be proven true.) In order to close this gap, we have created a Prolog predicate called `logicNot`, and we are developing rules to capture the semantics of logical negation. The `logicNot` predicate takes one argument, which must be a term: `logicNot(<term>)`. Three examples are presented in Table 4.

**Table 4. logicNot**

```
logicNot(isMemberOf(jones, sniper)).
logicNot(issubclassof(theaterObject, convoy)).
logicNot(logicNot(isclass(theaterObject))).
```

4.2. The Open World Assumption

In Prolog, the *closed world assumption* holds, which means that anything that cannot be proven true must be false. Alternatively, OWL has an *open world assumption*, meaning that a term is false only if it can be proven false. So, in Prolog, a term can only be true or false, while OWL also allows for the possibility that its truth value cannot be determined from the available information. This distinction is addressed with the `logicNot` predicate, which was presented in Section 4.1. If the user wants to ask a true/false question, Q, then it is necessary to submit two queries to SWORIER: `Q` and `logicNot(Q)`. Table 5 shows how the system's responses to these queries should be interpreted.

**Table 5. A True/False Query**

| `?- Q.` | `?- logicNot(Q).` | Is Q true? |
|---|---|---|
| yes | no | yes |
| no | yes | no |
| no | no | unknown |
| yes | yes | error |

4.3. Complementary and Disjoint Classes

Volz *et al.* (2003) claimed that "OWL features the `complementOf` primitive, which cannot be implemented in Horn Logics due to the fact, that there may be no negation in the head..." With the introduction of the `logicNot` predicate, this is no longer a problem. We can handle the complementary classes as well as the disjoint classes with the rules in Table 6.

**Table 6. Complementary and Disjoint Classes**

```
disjointClasses(C, D) :- complementaryClasses(C, D).
logicNot(isMemberOf(I, C)) :-
 disjointClasses(C, D), is_member_of(I, D).
isMemberOf(I, C) :-
 complementaryClasses(C, D), logicNot(is_member_of(I, D)).
```

4.4. Multiple Terms in the Head

One notable limitation of Horn rules is that the head (conclusion) of a rule cannot have more than one term. This means, if the conclusion of a rule is the conjunction of terms, it is necessary to create multiple rules. For example, the logical statement in Table 7a requires two rules in Prolog, as shown in Table 7b.

But for the logical rule in Table 7c, "...no Horn clause can be stated, since disjunction in the head would occur..." (Volz et al, 2003). We propose to address this problem by creating a new Prolog predicate that can be put in the head. So Table 7c can then be translated into Table 7d by using this new `or` predicate, which takes two arguments, each of which must be a term. Of course, we must supplement General Rules in order to properly establish the correct semantics of disjunction. Some examples of those rules are shown in Table 7e. Note that the last rule requires the `logicNot` predicate.

Unfortunately, the head of the last rule in Table 7e is a variable, which is not allowed in Prolog. However, although it may not be possible to solve this problem in general, because we are limiting our analysis to OWL, there are a finite number of predicates with which that variable can be instantiated, and this set of predicates does not require any knowledge of the particular ontologies or rules that are provided by the developer. So we can create one rule for each predicate, and some examples are presented in Table 7f.

**Table 7. Conjunction and Disjunction in the Head**

| | |
|---|---|
| a. | IF I is in C, THEN I is an individual, AND C is a class. |
| b. | `isindividual(I) :- isMemberOf(I, C).`<br>`isclass(C) :- isMemberOf(I, C).` |
| c. | IF C and D are complementary classes, AND I is an individual, THEN I is in C, OR I is in D. |
| d. | `or(ismemberof(I, C), ismemberof(I, D)):-`<br>`complementaryClasses(C, D), isIndividual(I).` |
| e. | `or(P, Q) :- P.`<br>`or(P, Q) :- or(Q, P).`<br>`or(or(P, Q), R) :- or(P, or(Q, R)).`<br>`Q :- or(P, Q), logicNot(P).` |
| f. | `isMemberOf(I, C) :- or(P, isMemberOf(I, C)), logicNot(P).`<br>`isSubClassOf(I, C) :- or(P, isSubClassOf(I, C)), logicNot(P).`<br>`logicNot(Q) :- or(P, logicNot(T)), logicNot(Q).`<br>`or(Q, R) :- or(P, or(Q, R)), logicNot(P).` |

4.5.Enumerated Classes

"The `owl:oneOf` primitive can be partially supported." (Volz et al, 2003) This primitive, which corresponds to our Prolog predicate, `isset`, defines a class, C, extensionally by providing a set of all and only the individuals in the class, $a_0$, ..., $a_n$. For example, Table 8a declares that there are exactly three members of the class `combatIntent`: `friendlyIntent`, `hostileIntent`, and `unknownIntent`.

**Table 8. Enumerated Class**

| | |
|---|---|
| a. | `isset(combatIntent, [friendlyIntent, hostileIntent, unknownIntent]).` |
| b. | `isMemberOf(a_i, C).` *for all i* |
| c. | `or(=(I, friendlyIntent),`<br>`or(=(I, hostileIntent), =(I, unknownIntent))) :- isMemberOf(I, combatIntent).` |

Volz *et al.* (2003) observed that Horn Logic rules could be developed that would draw the conclusions in Table 8b. But they also say that, "to support the other direction ... which states that every instance of C is one of the listed $a_i$ ... requires a disjunction in the consequent of the rule, which can not be provided by ... Horn Clauses." (Volz *et al.* 2003) However, with the `or` predicate, introduced in Section 4.4, this should no longer be a problem. Table 8c presents the rule that captures the semantics of the example in Table 8a.

4.6.Error Messages

It is desirable for SWORIER to test the data for consistency. For this purpose, we follow Volz *et al.* (2003) by implementing rules to catch inconsistencies, such as those in Table 9. (In order to check for errors, the developer must submit the query: `error(X).)` In each of these examples, the inconsistency is addressed by sending an error message to the developer. However, there are other ways to handle most inconsistencies. Examples will be presented in Sections 4.7 and 4.8.

**Table 9. Error Messages**

```
error(['A term cannot be both true and false.', P]) :- logicNot(P), P.
error(['The empty class cannot contain anything.', I]) :-  isMemberOf(I, nothing).
```

4.7. Existential Quantification

Prolog implicitly assumes that all variables in any rule are universally quantified. However, OWL can specify existentially quantified variables. For example, the OWL code in Table 10a states that every theater object is described by at least one observation artifact. There are three ways to enforce this restriction. The simplest is to report an error message if it is violated, as in Section 4.6. Volz *et al.* (2003) uses the technique of skolemization. And the third approach is to add new facts to the knowledge base, which is discussed in Section 4.8.

Skolemization solves the problem of a violated existential restriction by letting a specific term represent the missing individual. This term must be unambiguous, so its argument variables are selected to make it distinct. For example, given the restriction in Table 10a, if there is a theater object, I, that is not described by any observation artifacts, then the term `unnamedIndividual(I, describedBy, observationArtifact)` is used to represent the required observation artifact. In general, the two rules in Table 10b are applied in order to insure that this term satisfies the existential restriction.

**Table 10. An Existential Constraint**

| | |
|---|---|
| a. | <owl:Class rdf:about="#theaterObject" /><br><rdfs:subClassOf><br><owl:Restriction><br><owl:onProperty rdf:resource="#describedBy" /><br><owl:someValuesFrom rdf:resource="#observationArtifact" /><br></owl:Restriction><br></rdfs:subClassOf><br></owl:Class> |
| b. | haspropertywith(I, P, unnamedIndividual(I, P, C2)) :-<br>hassomevaluesofpropertyfrom(C1, P, C2), isMemberOf(I, C1).<br>ismemberof(unnamedIndividual(I, P, C2), C2) :-<br>hassomevaluesofpropertyfrom(C1, P, C2), isMemberOf(I, C1). |

<u>4.8. Cardinality</u>

In OWL, there are three cardinality primitives: (1) `minCardinality`, (2) `maxCardinality`, and (3) `cardinality`. Each of these primitives takes three arguments: a class, a property, and a number. The primitives' meanings are that each individual in the given class participates in the given property with (1) at least, (2) at most, or (3) exactly the given number of unique individuals.

**Table 11. Cardinality Rules**

| | |
|---|---|
| a. | <owl:Class rdf:ID="theaterObject" /><br><rdfs:subClassOf><br><owl:Restriction><br><owl:onProperty rdf:resource="#describedBy" /><br><owl:cardinality rdf:datatype="&xsd;nonNegativeInteger"><br>1<br></owl:cardinality><br></owl:Restriction><br></rdfs:subClassOf><br></owl:Class> |
| b. | equivalentindividuals(I1, I2) :- isMemberOf(I, theaterobject),<br>hasPropertyWith(I, describedby, I1),<br>hasPropertyWith(I, describedby, I2). |
| c. | enforceConstraints :- isMemberOf(I1, theaterobject),<br>not(hasPropertyWith(I1, describedby, I2)),<br>gensym(newIndividual, I3),<br>assert(hasPropertyWith(I1, describedby, I3)). |

Volz *et al.* (2003) claims that, "the unrestricted use of cardinality constraints cannot be supported efficiently in Logic Programming environments..." It is true that, in theory, there are an unlimited number of cardinality constraints that the developer could impose. However, we can extend SWORIER's system design (from Figure 2) by introducing a new module, as shown in Figure 3. Any cardinality constraints found in the ontologies are sent to this *Cardinality Rules* module, which produces one or two Prolog rules for each constraint. Since there are a finite number of cardinality constraints in any ontologies, it is possible to develop all and only the necessary cardinality rules, and thus, the problem is tractable.

An example cardinality constraint is expressed by the OWL code in Table 11a, which says that every theater object is described by exactly one individual. For this constraint, it is necessary to generate two rules, one testing to make sure that there is no more than one individual, and the other checking that there is at least one individual that satisfies the restriction. We have two options for the first rule: 1) If two different individuals are found that both describe the same theater object, we could report an error to the developer, as in Section 4.6, or 2) we could enforce the constraint with the rule in which says that if any theater object is described by two individuals, then those two individuals must be equivalent. For the second rule, there are three options: 1) If a theater object exists that is not described by any individuals, we could report an error to the developer, as in Section 4.6, 2) the constraint could be enforced by skolemization, which was explained in Section 4.7, or 3) the problem could be fixed by adding new facts to the knowledge base. The last option is demonstrated in Table 11c, where the Prolog predicate, `gensym`, sets I3 to a new unique constant, and the `assert` predicate adds the required fact to the knowledge base. (The query, `enforceConstraints`, would be run offline in order to create all of the required facts.)

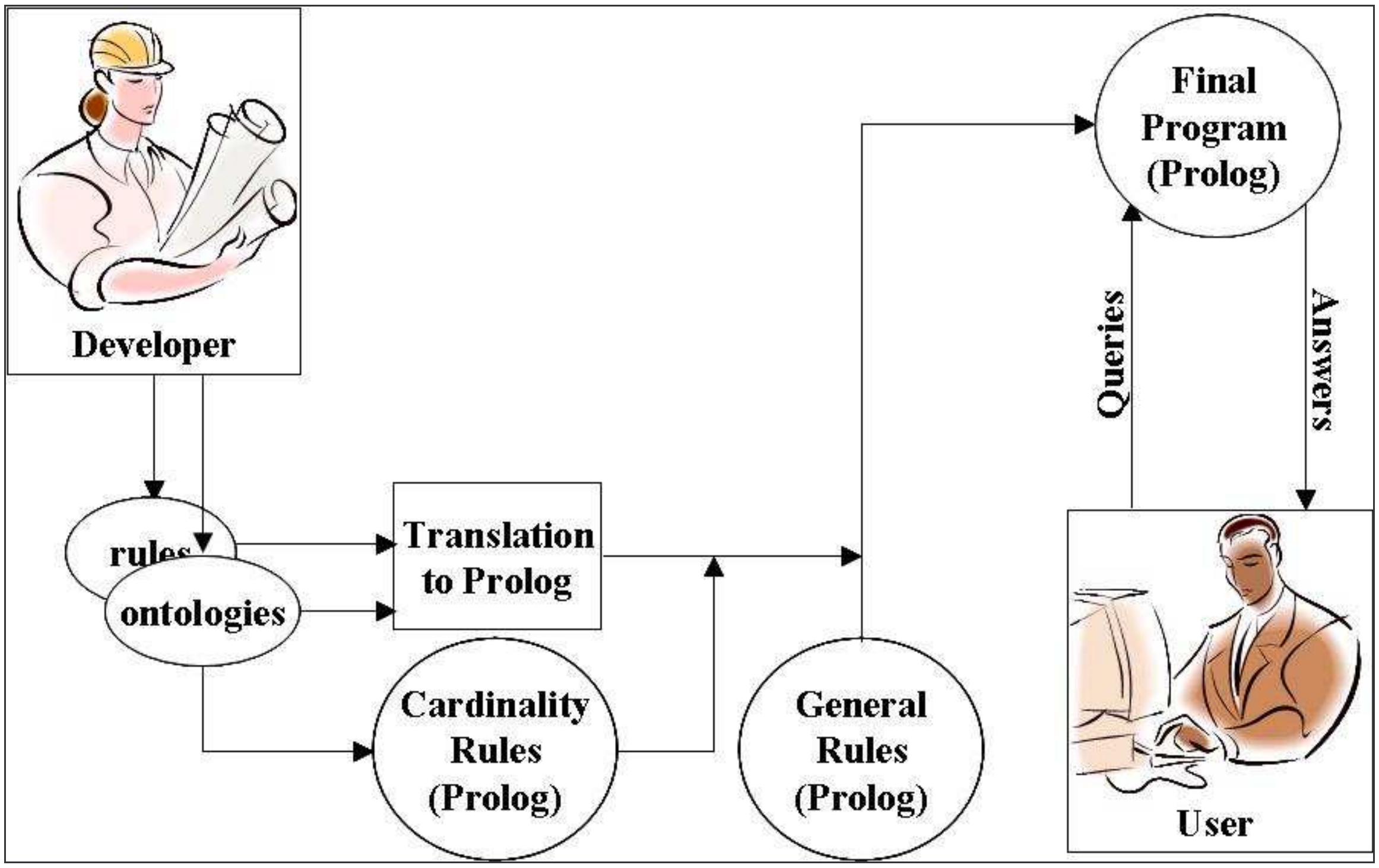

**Figure 3. The Cardinality Rules Module**

4.9.Duplicate Facts

It is desirable to prevent SWORIER from generating the same fact more than once. To demonstrate the rationale, consider the program in the Table 12a. The query, `fact(X, c, 1)` would cause the system to return two copies of `fact(a, c, 1)`; one because of rule 1, and the other through the interaction between rules 2 and 3. Although it is not difficult to remove repetitive facts in a post-processing procedure, a significant cost in efficiency can still result. Consider, for example, the query `fact3(a, c, 3)`. To determine the answer, the system tests the first term in the body of line 3, `fact(a, c, 1)`, and it succeeds twice. Then the system runs two tests on the second term, `fact(a, c, 1)`, each time finding two results. Thus, the third term, `slow(c)`, must be tested four times, unnecessarily quadrupling the time spent processing that term. And for the query `fact3(a, c, 5)`, the `slow(c)`test is run 16 times. So it should be clear that, when duplicate facts are generated, they can potentially slow down the program significantly.

To block duplicate facts, we can add the term, `not(Y=c)` to rule 2, as shown in Table 12b. This prevents `fact(a, c, 1)` from being generated via rule 2. But given the query `fact(a, Y, 2)`, the system fails to return `fact(a, b, 2)`, which should be proven by rules 2 and 3. This is because, in order to test `not(Y=c)`, Prolog tries to prove `Y=c`. But this is easy, since `Y` is an unbound variable, so it can be set to `c`. This causes `Y=c` to succeed, and so `not(Y=c)` fails, and the rule is incorrectly blocked.

By changing rule 2 as shown in Table 12c, we can insure that Y is bound before `not(Y=c)` is tested. Now the program works correctly, and the duplicates are blocked. But unfortunately, the system must investigate the `fact(X, Y, N-1)` term, even if the block, `not(Y=c)`, is doomed to fail. In addition, rule 2 is no longer tail recursive, so the Prolog compiler cannot utilize a significant efficiency improvement.

**Table 12. Duplicate Facts**

| | |
|---|---|
| a. | 1. `fact(a, c, N) :- N>=0.`<br>2. `fact(X, Y, N) :- N>0, fact(X, Y, N-1).`<br>3. `fact3(X, Y, N) :- fact(a, Y, N-2), fact(X, c, N-2), slow(Y).` |
| b. | 1. `fact(a, c, N) :- N>=0.`<br>2. `fact(X, Y, N) :- not(Y=c), fact(X, Y, N-1).`<br>3. `fact(X, b, 1).` |
| c. | 1. `fact(a, c, N) :- N>=0.`<br>2. `fact(X, Y, N) :- N>0, fact(X, Y, N-1), not(Y=c).` |
| d. | 1. `fact(a, c, N) :- N>=0.`<br>2. `fact(X, Y, N) :- isLetter(Y), not(Y=c), N>0, fact(X, Y, N-1).` |

We believe we can make these rules both correct and efficient. The key is that we require all constants to be declared with predicates like `isclass`, `isindividual`, `isproperty`, and `isdatatype`. Then, by specifying the required type of `Y` at the beginning of the rule's body, as in rule 2 in Table 12d, this has the desired effect of binding `Y`, enabling the blocker `not(Y=c)` to be tested.

4.10.Cyclic Hierarchies

Cyclic class hierarchies and cyclic property hierarchies can be problematic. Suppose the given OWL ontology includes the facts shown in Table 13a. Then the computation of the transitive closure of subclass, using the rules from Table 13b, produces an infinite number of responses to the query, `isSubClassOf(X, Y)`, as the system loops around and around the cycle. Even though we may claim that a cyclic hierarchy is erroneous, we cannot prevent the developer from creating one. So SWORIER should be able to handle it.

We propose changing the subclass transitive closure rules (Table 3a) into the rules in Table 13b. The idea is to stop the cycle when it reaches the beginning again, which occurs when the two parameters of `isSubClassOf` are equal. For this purpose, we create a new predicate `is_sub_class_of_but_not_equal_to` that includes all of the subclass relations, except for the reflexive ones. (The first rule catches them.) Note that we use the technique discussed in Section 4.9, by including `isclass` predicates to insure that the variables are bound before running any `not` tests on them.

**Table 13. Cyclic Hierarchies**

| | |
|---|---|
| a. | `issubclassof(armedForce, coalition).`<br>`issubclassof(coalition, politicalGroup).`<br>`issubclassof(politicalGroup, armedForce).` |
| b. | `isSubClassOf(C, C).`<br>`isSubClassOf(C, D) :- is_sub_class_of_but_not_equal_to(C, D).`<br>`is_sub_class_of_but_not_equal_to(C, D) :- issubclassof(C, D).`<br>`is_sub_class_of_but_not_equal_to(C, E) :-`<br>`isclass(C), isclass(E), not(C=E),`<br>`issubclassof(C, D), is_sub_class_of_but_not_equal_to(D, E).` |

4.11. Anonymous Classes

OWL can define classes called anonymous classes without actually naming them. Table 14a has an example of an anonymous class, and Table 14b has our suggestion of how to translate it. An anonymous class, `unnamedClass(hasCombatIntent, friendly-Intent)`, is generated like anonymous individuals that were presented in Section 4.7.

**Table 14. Anonymous Classes and Properties**

| | |
|---|---|
| a. | `<owl:Class>`<br>`<hasCombatIntent rdf:resource="#friendlyIntent"/>`<br>`</owl:Class>` |
| b. | `hasallvaluesofpropertyfrom(`<br>`unnamedClass(hasCombatIntent, friendlyIntent),`<br>`hasCombatIntent,`<br>`friendlyIntent).` |

## 5. Dynamic Changes

Another useful capability is to change the knowledge base at run time. For example, in our convoy task, intelligence reports can come in at any time during a scenario, and we want SWORIER to be able to incorporate the new information into the knowledge base. This must be done within a few seconds. So, we have enabled SWORIER to accommodate *dynamic changes* of facts, adding or removing facts at run time. (We have not yet tried dynamically changing classes, properties, or rules.) Unfortunately, dynamic assertions significantly decrease efficiency, because the Prolog compiler can no longer be used. It is not possible to assert or retract any facts with predicates that are compiled. This issue is addressed in Section 6.3.

SWORIER is also capable of dynamically changing rules, but only in a restricted way. We require that all of the desired rule sets are available in advance. This can still be quite useful. For example, under low visibility conditions, different rules might be desired from the rules used with high visibility. Both rule sets can be developed in advance, and then SWORIER can generate a separate program for each case. These rules might be considered different policies or rules invoked by different contexts. At run time, when visibility is high, the user queries are submitted to the first program. But whenever visibility is lost, such as at the onset of a sandstorm, the two programs are swapped, and the user queries are sent to the second program.

Figure 4 shows the system after it is extended to accommodate dynamic changes. Note that whenever facts are added or removed from the knowledge base, both programs must be modified appropriately.

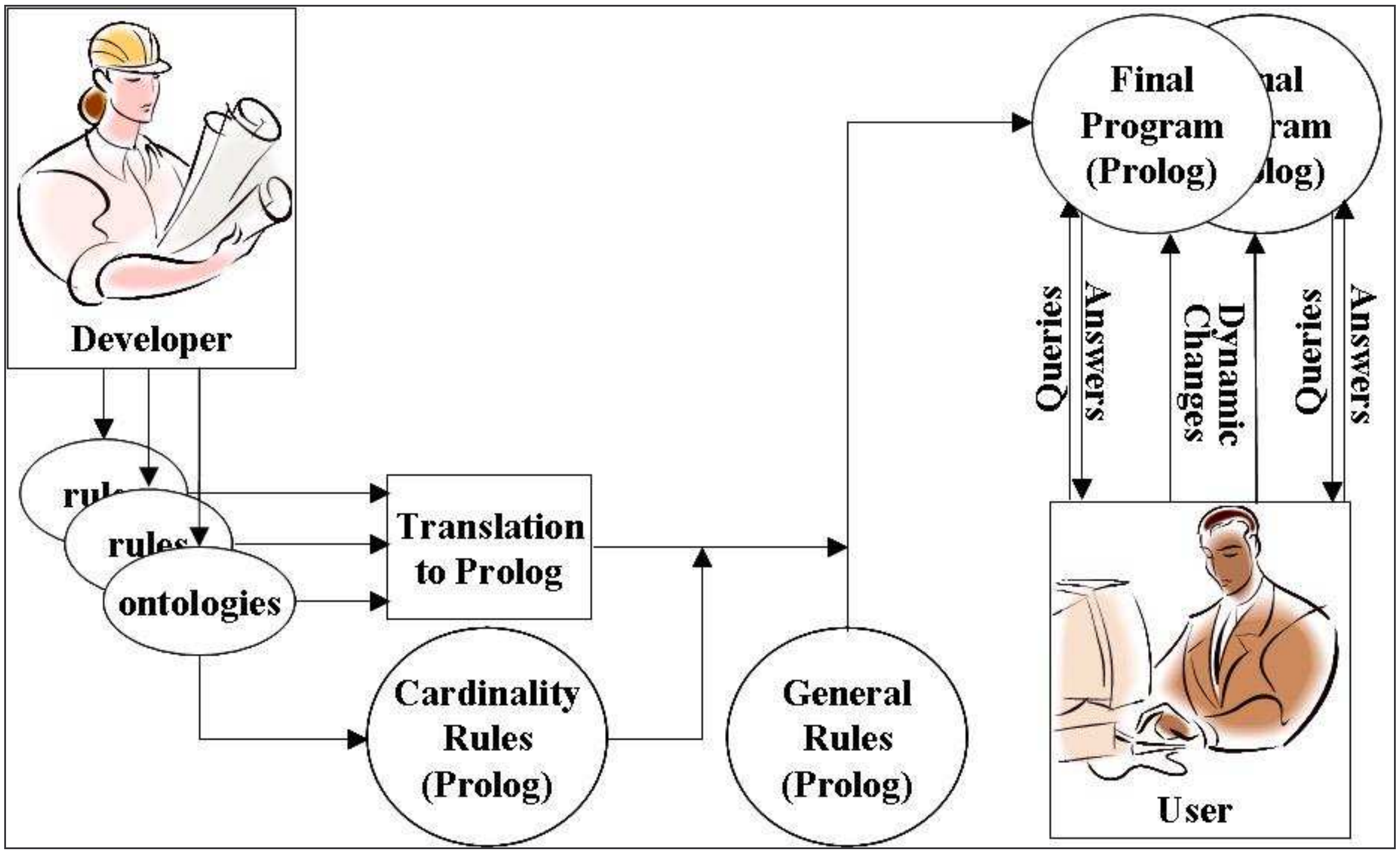

**Figure 4. Dynamic Changes of Facts and Rules**

## 6. Efficiency

Initially, the SWORIER system was too slow. As shown in Table 15a, it took 1.9 hours to incorporate two dynamic changes into the knowledge base: A report of the convoy's current position and speed and information about a motorized infantry unit approaching it from ahead. After those changes were made, 1.5 hours were needed to respond to the following queries: What are the positions and speeds of all known units, and what are the current alerts and recommendations?

**Table 15. Response Time (online)**

| | Extensionalization | Code Minimization | Dynamic Changes | Queries |
|---|---|---|---|---|
| a. | no | no | 1.9 hours | 1.5 hours |
| b. | yes | no | 25.2 minutes | 58 minutes |
| c. | yes | yes | 10 milliseconds | 130 milliseconds |

The time efficiency that is required depends on the application. For our military task, once a mission begins, the system's responses must be very fast. If it takes more than a few seconds to answer a query at run time, the system is effectively useless. However, before the mission begins, more time is generally available for knowledge compilation. Still, this offline processing would usually need to be done in hours, not days.

6.1. Extensionalization

In order to make the system tractable at run time, we implemented an offline technique to speed up the program. We modified SWORIER to extensionalize all of the facts that can be derived from the input (that a user might want to query on), converting the program from an intensional form to an extensional form. Figure 5 shows the modified system design.

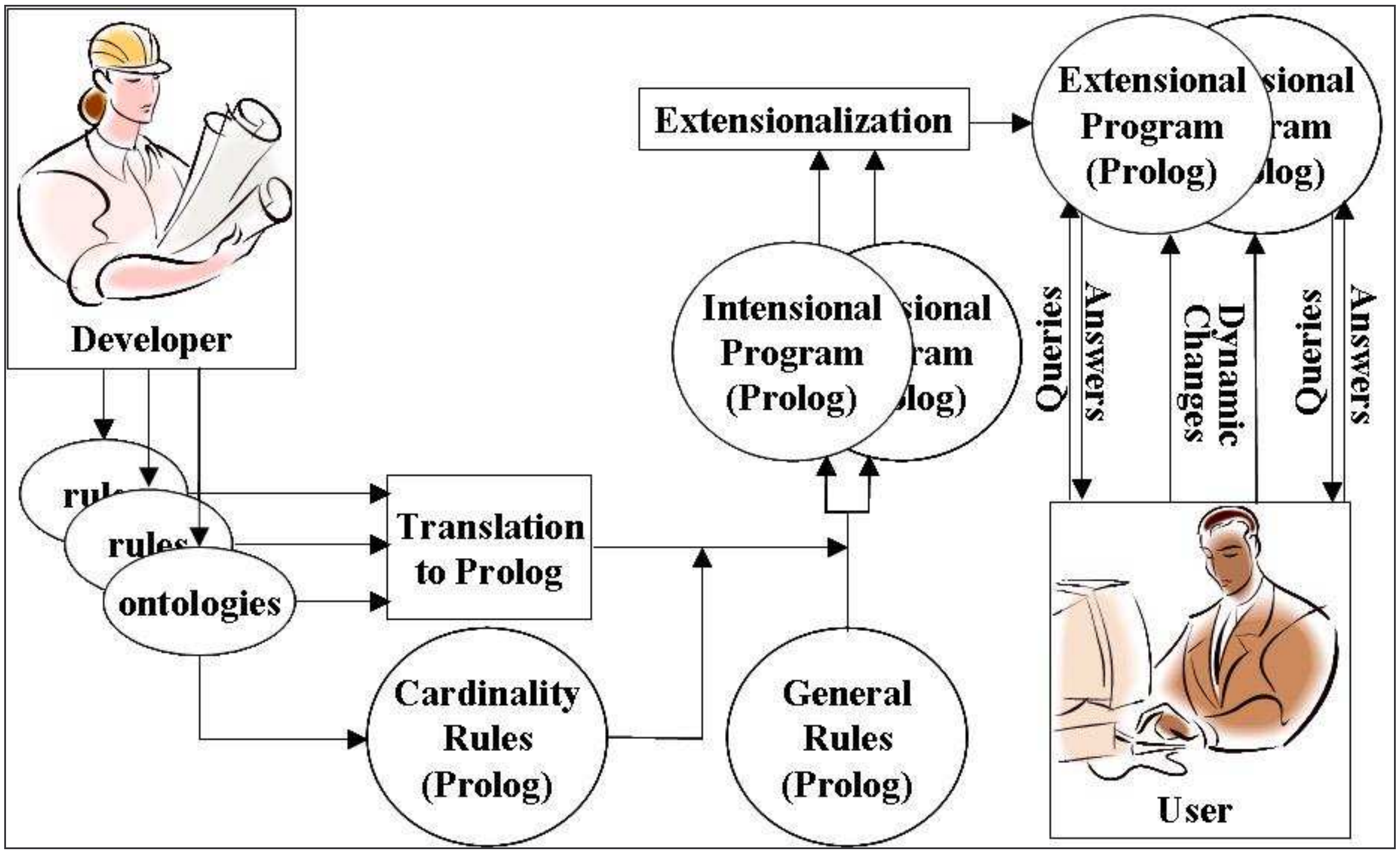

**Figure 5. Extensionalization**

Table 16a shows some sample intensional code, and the corresponding extensional code can be found in Table 16b. The extensionalization algorithm runs generalized queries (`isClass(C)` and `equivalentClasses(C, D)` in order to derive all of the desired facts and save them in the extensional program. (Note that it is easy to keep duplicate facts out of the extensional code, because they always look identical.) Then, at run time, the extensional program uses the derived facts. Since it only consists of facts in camelcase form, several of the rules do not apply to the extensional code. Only the facts that are added dynamically to the knowledge base have all-lowercase predicates.

**Table 16. Intensional and Extensional Code**

| | |
|---|---|
| a. | `isclass(regionOfInterest).`<br>`isclass(roi).`<br>`equivalentclasses(regionOfInterest, roi).`<br>`isClass(C) :- isclass(C).`<br>`equivalentClasses(C, D) :- equivalentclasses(C, D).`<br>`equivalentclasses(C, C) :- isClass(C).`<br>`equivalentClasses(D, C) :- equivalentclasses(C, D).` |
| b. | `isClass(regionOfInterest).`<br>`isClass(roi).`<br>`equivalentClasses(regionOfInterest, roi).`<br>`equivalentClasses(regionOfInterest, regionOfInterest).`<br>`equivalentClasses(roi, roi).`<br>`equivalentClasses(roi, regionOfInterest).` |

This preprocessing technique enabled the system to work much faster, as shown in Table 15b. However, it still required 25.2 minutes to incorporate the same two dynamic changes as in the previous test, and to answer the two queries took 58 minutes. This is still unacceptably slow. In addition, the offline extensionalization process caused the AMZI Prolog application to crash, as shown in Table 17a. We presume that the computer ran out of memory.

**Table 17. Extensionalization Time (offline)**

| | Avoiding Reanalysis | Code Minimization | Extensionalization |
|---|---|---|---|
| a. | no | no | CRASH |
| b. | yes | no | 13 hours |
| c. | yes | yes | 6.5 hours |

6.2. Avoiding Reanalysis

In the process of extensionalizing the code, it was very common to test a term several times with the same arguments. This unnecessary processing can be very slow. For example, given the code in Table 18, the system must test `isSubClassOf(convoy, theaterobject)` at least twice: Once when searching for all of the true `isSubClassOf` terms, and again when trying to prove `isMemberOf(convoy1, theaterobject)`.

**Table 18. Reevaluating a Term**

```
ismemberof(convoy1, convoy).
issubclassof(convoy, militaryunit).
issubclassof(militaryunit, theaterobject).
isSubClassOf(C, D) :- issubclassof(C, D).
isSubClassOf(C, E) :- issubclassof(C, D), isSubClassOf(D, E).
isMemberOf(I, C) :- ismemberof(I, C).
isMemberOf(I, D) :- isSubClassOf(C, D), isMemberOf(I, C).
```

The proof of `isSubClassOf(convoy, theaterobject)` takes five steps.[3] In general, a very slow test may be run several times. To avoid the reevaluation of a term, each time an `isSubClassOf` term is tested, that term is asserted as a success or failure. Then, the next time the term needs to be tested, the answer is found in the new assertion, so it is not necessary to run the full test again.

[3]
1. `isSubClassOf(convoy, theaterobject) :-`
   `issubclassof(convoy, theaterobject).` *(FAILS)*
2. `isSubClassOf(convoy, theaterobject) :-`
   `issubclassof(convoy, D),`
   `isSubClassOf(D, theaterobject).`
3. `issubclassof(convoy, militaryunit).`
4. `isSubClassOf(militaryunit, theaterobject) :-`

**Table 19. The Code Minimization Algorithm**

| | | | |
|---|---|---|---|
| a. | **Base Case 1:** | IF<br>and<br>THEN | P is a built-in predicate (reserved keyword) in Prolog,<br>P is not `findall`,<br>P is a satisfiable predicate. |
| | **Base Case 2:** | IF<br>THEN | P is a predicate in a fact,<br>P is a satisfiable predicate. |
| | **Base Case 3:** | IF<br>and<br>THEN<br>and | P is the predicate in the head of a rule, R,<br>each predicate in R's body is P,<br>P is a satisfiable predicate,<br>R is a satisfiable rule. |
| | **Inductive Case 1:** | IF<br>and<br>THEN<br>and | P is the predicate in the head of a rule, R,<br>every predicate in R's body is satisfiable,<br>P is a satisfiable predicate,<br>R is a satisfiable rule. |
| | **Inductive Case 2:** | IF<br><br>and<br>and<br><br>THEN | a rule, R, has a term, T, in its body, where T is `assert(F)`, `asserta(F)`, or `assertz(F)`,<br>P is the predicate of F,<br>all of the predicates preceding T in the body of R are satisfiable predicates,<br>P is a satisfiable predicate. |
| | **Inductive Case 3:** | IF<br>and<br>THEN | A is the second argument of a `findall`,<br>all of the predicates in A are satisfiable predicates,<br>the `findall` is treated as if it was a satisfiable predicate |
| b. | **Base Case:** | IF<br><br>THEN | P is a predicate that can be called from outside the rules, [4]<br>P is a testable predicate. |
| | **Inductive Case 1:** | IF<br>THEN | `findall` is determined to be a testable predicate,<br>each predicate in P's second argument is also testable. |
| | **Inductive Case 2:** | IF<br><br>and<br><br>THEN<br>and | P is a predicate such that P or `not(P)` is found in the body of a rule, R,<br>R's head can be determined to have a testable predicate,<br>P is a testable predicate,<br>R is a testable rule. |

```
         issubclassof(militaryunit, theaterobject).
   5. issubclassof(militaryunit, theaterobject).
```

[4] These are the predicates in queries, dynamically asserted facts, and the call to initiate extensionalization.

Avoiding reevaluation of `isSubClassOf` makes the offline extensionalization process tractable, though it still takes more than 13 hours, as shown in Table 17b. We also tried implementing the avoiding reevaluation technique on `isMemberOf`. But, for our task, very few `ismemberof` facts are known until runtime, and the cost of overhead outweighs the benefit, making extensionalization slower.

### 6.3.Minimizing Code

Another efficiency improvement can be implemented if certain knowledge is available prior to run time. Given a list of all of the predicates that are used in 1) the ontology, 2) the dynamic changes, and 3) the queries, it may be possible to eliminate some of the rules, thus improving efficiency of the extensionalization process. In addition, the same technique can be used to eliminate rules in the run time program.

The idea is to figure out which of the rules are actually necessary, because the unnecessary rules can be dropped, improving efficiency. If a rule can never successfully fire, then it is unnecessary. Also, if no query will ever result in testing a rule, then that rule is unnecessary.

More precisely, a rule is a *necessary rule* only if it is both *satisfiable* and *testable*. To determine which rules are satisfiable and testable, it is necessary to figure out which predicates are satisfiable and testable, respectively. The algorithm that defines satisfiable rules and predicates is presented in Table 19a, and Table 19b shows how to determine which rules and predicates are testable. (In addition, we were able to drop more rules by assuming that the knowledge base was already consistent, eliminating the need to test for consistency.)

After removing all of the unnecessary rules, the extensionalization process only took 6.5 hours, as shown in Table 17c. And the online processes can run much faster, because the Prolog compiler can be applied to all of the predicates that are not changed dynamically. Table 15c shows that it requires only 10 milliseconds to assimilate the two dynamic changes and 130 milliseconds to answer the two queries. (The dynamic changes and queries used in our experiments are briefly described near the beginning of Section 6.) These results satisfy the requirements of our military task.

## 7. Related Work

Recent research has addressed similar issues and problems concerning the interaction of Semantic Web ontology and rule technologies and logic programming. Related work includes research on answer set programming (Eiter *et al.* 2004; Heymans & Vermeir 2003), disjunctive logic programming (Maedche & Volz 2003; Minker & Seipel 2002), constructive negation (Barták & Roman 1998), and Description Logic Programming (DLP) (Grosof 2003). However, we have not yet had an opportunity to investigate this other work enough to intelligently comment on it. Our preliminary experimentation with answer set programming, however, seems to demonstrate across-the-board gains in efficiency, compared to our Prolog implementation. But this work is as yet incomplete.

## 8. Discussion

The SWORIER system, given ontologies and rules, serves as an engine that responds to queries. This type of work will result in a significant enhancement to the Semantic Web, by providing a generally useful service to any application that requires information from the Semantic Web. SWORIER is also amenable to dynamic changes, quickly assimilating new facts or retracting old facts in an operational setting. Although it cannot handle dynamic additions, deletions, or modifications of rules at this time, it can switch between predefined sets of rules on the fly. Previously there has been little work involving rules and dynamic changes.

We have built our work on the foundation developed in a paper written by Voltz *et al.* (2003). We have addressed five of the problems that this paper suggested were unsolvable. 1) By defining logical negation in Prolog, which makes it possible to satisfy the open world assumption, it is now possible to properly capture the semantics of complementary classes, as well as disjoint classes. 2) Disjunction in the head of a rule is captured with the use of a new disjunctive operator. 3) The disjunctive operator enables the analysis of enumerated classes. 4) Through an offline analysis of the given ontologies, SWORIER can automatically develop rules that enforce all and only the given cardinality constraints. 5) Using a different syntax to express OWL facts in Prolog, addressing properties of equivalent individuals has been simplified. Also, we dealt with three other issues: duplicate facts, cyclical hierarchies, and anonymous classes.

We have also introduced alternative approaches for dealing with inconsistencies in the given information. When an inconsistency is discovered, it is always possible to simply send an error message to the developer. However, for many inconsistencies, the possibility of fixing the problem automatically is available. And, with existential constraints and minimum cardinality constraints, this can be done by adding new facts to the knowledge base or by applying the skolemization method. We have not yet confirmed which of these alternatives is preferable in which situations. We expect that by introducing pragmas (instructions/annotations to the knowledge compilation process), we can allow the developer to choose the specific behavior he/she wants.

Efficiency problems have been addressed through 1) extensionalization, which is a tabling method that converts a set of rules and facts into a set of facts, 2) avoiding reanalysis, which saves results the first time they are determined to avoid running the same costly evaluation again, and 3) code minimization, which deletes rules that are unnecessary, for both offline and online processing. In our experiments, the offline compilation process now completes in 6.5 hours, two dynamic changes are incorporated into the knowledge base in 10 milliseconds, and two queries can be answered in only 130 milliseconds.

In the future, we hope to analyze and convert the developer's rules into appropriately optimized rules, perhaps via the use of rewrite rules as for magic sets optimization and related rule techniques for logic queries (Cadoli *et al.* 1999; Sippu & Soisalon-Soininen 1996). (See Section 3.3.) Also, the ideas presented in Section 4 concerning multiple terms in the head, equivalent individuals, cardinality issues, cyclic hierachies, and anonymous classes have not yet been implemented. In addition, there are several primitives in OWL that are not yet implemented in SWORIER, including: cardinality, or, and subPropertyOf,[5] and the logicNot predicate, introduced in Section 4.1 is only partially implemented.

**Acknowledgments**

The authors' affiliation with The MITRE Corporation is provided for identification purposes only, and is not intended to convey or imply MITRE's concurrence with, or support for, the positions, opinions or viewpoints expressed by the authors. We note that the views expressed in this paper are those of the authors alone and do not reflect the official policy or position of any other organization or individual.

---

[5] The other OWL primitives that are not yet implemented: `AllDifferent`, `DatatypeProperty`, `differentFrom`, `distinctMembers`, `domain`, `equivalentIndividuals`, `equivalent-Properties`, `FunctionalProperty`, `hasValue`, `InverseFunctionalProperty`, `max-Cardinality`, `minCardinality`, `range`, `sameAs`, `SymmetricProperty`, and `Transitive-Property`.